\documentclass{bmvc2k}

\usepackage[utf8]{inputenc}
\usepackage{soul}
\usepackage{pifont}
\usepackage{amsmath}
\usepackage{amssymb}
\usepackage{tabularx}
\usepackage{capt-of}
\usepackage{booktabs}
\usepackage{algorithm}
\usepackage[noend]{algpseudocode}
\usepackage{enumitem}

\usepackage[capitalize,noabbrev]{cleveref}

\newcommand{\cmark}{\ding{51}}%
\newcommand{\argmin}{\operatornamewithlimits{argmin}}

\newcommand{\fapp}[2]{\ensuremath{#1}\tiny{$\pm$ \ensuremath{#2}}}    
\renewcommand{\fapp}[2]{\ensuremath{#1}}

\newcommand{\fappb}[2]{\fapp{\textbf{#1}}{#2}}
\newcommand{\faau}[2]{\fapp{\underline{#1}}{#2}}

\newcommand{\dpp}{DER\texttt{++}\xspace}
\newcommand{\classil}{ClassIL\xspace}
\newcommand{\ie}{\textit{i.e.}\,\,}
\newcommand{\ienospace}{\textit{i.e.}}
\makeatletter
\@ifundefined{c@rownum}{%
  \let\c@rownum\rownum
}{}
\@ifundefined{therownum}{%
  \def\therownum{\@arabic\rownum}%
}{}
\makeatother

\newcommand{\quotationmarks}[1]{``#1''}
\newcommand{\tit}[1]{\smallbreak\noindent\textbf{#1 }}

\Crefname{equation}{Eq.}{Eqs.}
\Crefname{figure}{Fig.}{Figs.}
\Crefname{table}{Tab.}{Tabs.}
\Crefname{tabular}{Tab.}{Tabs.}
\Crefname{section}{Sec.}{Secs.}
\Crefname{algorithm}{Algo.}{Algs.}
\usepackage{multirow}

\title{May the Forgetting Be with You: Alternate Replay for Learning with Noisy Labels}

\addauthor{Monica Millunzi}{monica.millunzi@unimore.it}{1,2}
\addauthor{Lorenzo Bonicelli}{lorenzo.bonicelli@unimore.it}{1}
\addauthor{Angelo Porrello}{angelo.porrello@unimore.it}{1}
\addauthor{Jacopo Credi}{jacopo.credi@axyon.ai}{2}
\addauthor{Petter N. Kolm}{petter.kolm@nyu.edu}{3}
\addauthor{Simone Calderara}{simone.calderara@unimore.it}{1}

\addinstitution{
 AImageLab\\
 University of Modena and Reggio Emilia\\
 Modena, Italy
}
\addinstitution{
 AxyonAI\\
 Viale dell'Autodromo, 210\\
 Modena, Italy
}
\addinstitution{
 Courant Institute of Mathematical Science\\
 New York University (NYU)\\
 New York, USA
 }

\runninghead{Millunzi, et al.}{Alternate Replay for Learning with Noisy Labels}

\def\eg{\emph{e.g}\bmvaOneDot}

\begin{document}

\maketitle


\begin{abstract}
Forgetting presents a significant challenge during incremental training, making it particularly demanding for contemporary AI systems to assimilate new knowledge in streaming data environments. To address this issue, most approaches in Continual Learning (CL) rely on the replay of a restricted buffer of past data. However, the presence of noise in real-world scenarios, where human annotation is constrained by time limitations or where data is automatically gathered from the web, frequently renders these strategies vulnerable. In this study, we address the problem of CL under Noisy Labels (CLN) by introducing Alternate Experience Replay (AER), which \textit{takes advantage of forgetting} to maintain a clear distinction between clean, complex, and noisy samples in the memory buffer. The idea is that complex or mislabeled examples, which hardly fit the previously learned data distribution, are most likely to be forgotten. To grasp the benefits of such a separation, we equip AER with Asymmetric Balanced Sampling (ABS): a new sample selection strategy that prioritizes purity on the current task while retaining relevant samples from the past. 
Through extensive computational comparisons, we demonstrate the effectiveness of our approach in terms of both accuracy and purity of the obtained buffer, resulting in a remarkable average gain of $4.71\%$ points in accuracy with respect to existing loss-based purification strategies. Code is available at \url{https://github.com/aimagelab/mammoth}.
\end{abstract}
\section{Introduction}
\label{sec:intro}
Despite the latest breakthroughs, modern AI still struggles to learn in a continuous fashion and suffers from \textit{catastrophic forgetting}~\cite{mccloskey1989catastrophic}, \ie the new knowledge quickly replaces all past progress. Therefore, Continual Learning (CL) has recently gathered an increasing amount of attention: among the others, one prominent strategy is to interleave examples from the current and old tasks (rehearsal). To do so, a small selection of past data is retained in a memory buffer~\cite{van2019three,chaudhry2019tiny}, as in Experience Replay (ER)~\cite{ratcliff1990connectionist,robins1995catastrophic}.
\begin{figure}[t]
    \centering
    \includegraphics[width=0.95\textwidth]{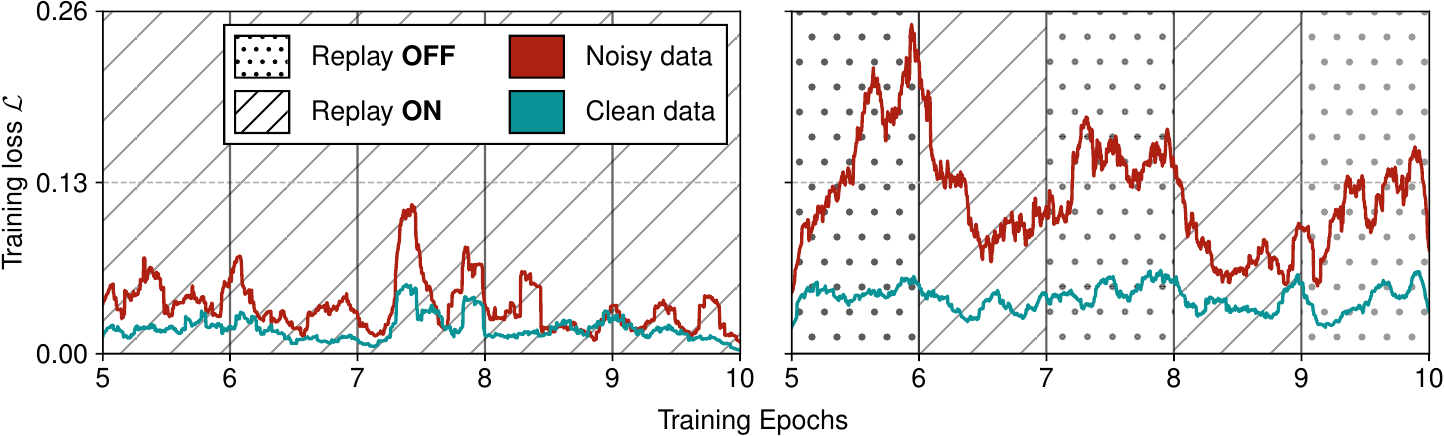}
    \caption{Training loss of clean and noisy during the second task of Seq.\ CIFAR-10 with $40\%$ noise. The loss is computed on examples from the first task stored in the memory buffer. Standard replay makes the two indistinguishable (\textit{left}) but alternating epochs of replay and forgetting maintain a significant loss separation (\textit{right}).}
    \label{fig:buffer_loss_divergence}
\end{figure}

Intuitively, the effectiveness of these methods depends strictly on the memory content: the larger the gap between the memory and the true distribution underlying all the previous tasks, the lower the chances of learning a reliable model. In this respect, several factors may intervene and degrade the snapshot portrayed by the buffer. Several works have highlighted the shortcomings of low-capacity buffers and their link to severe overfitting~\cite{verwimp2021rehearsal,bonicelli2022effectiveness}. More recently, the plausible presence of annotation errors has emerged as an engaging factor~\cite{kim2021continual,bang2022online}, due to the subsequent poisoning the memory buffer would be subjected to. Indeed, not only would a few observations of past tasks be available for the learner, but they might be erroneously annotated. It is noted that the presence of \textit{noisy annotations}~\cite{Xiao_2015_CVPR,lee2018cleannet,li2017webvision,2018IRCDL} is an inescapable characteristic of CL: to allow the learner to digest incoming training examples on-the-fly, data has to be annotated within a restricted temporal window, leading to poor human annotations and hasty quality controls. In light of this, preliminary works~\cite{kim2021continual,bang2022online} focus on purifying the memory buffer and then consolidate their knowledge at the end of the current task (or while learning the task itself~\cite{karim2022cnll}). To do so, they spot clean samples by leveraging the popular \textit{small-loss criterion}~\cite{arpit2017closer,han2018coteaching,jiang2018mentornet} and the fact that the most trustworthy examples are those favoured during the first training stages (\textit{memorization effect}), thus exhibit a lower value of the loss function. However, despite its effectiveness in the offline scenario, such a criterion may be weak in incremental settings. Indeed, as learning does not re-start from scratch but builds upon previous knowledge, the adaption is faster and hence the loss-value separation between clean and noisy samples tends to vanish~\cite{zhang2016understanding,arpit2017closer}.

To overcome this limitation, we explore a radically different approach which could be summarized by a quote ascribed to Julius Caesar \quotationmarks{if you can't defeat your enemy, have him as your friend}: while existing methods see forgetting only as an issue to solve, we use it to identify noisy examples within the data stream. We build upon the work of~\cite{toneva2018an,maini2022characterizing}, which theoretically demonstrates that mislabeled examples are quickly forgotten, whereas complex or rare instances tend to be retained for longer periods or may not be forgotten at all.

To illustrate such a phenomenon, we depict the loss trend of clean and noisy samples in a memory buffer produced by a rehearsal baseline (ER-ACE~\cite{caccia2022new}). In particular,~\cref{fig:buffer_loss_divergence} (left) shows the loss value sampled during standard training; differently, in~\cref{fig:buffer_loss_divergence} (right) we alternatively switch replay regularization on and off at each epoch. As can be seen, stopping replay has a distinct impact: while the loss value of clean samples remains low, it hugely increases for mislabeled ones. We remark that this gap holds even when replay regularization turns on, as the model easily adapts to clean samples and hence learns them faster~\cite{arpit2017closer,jiang2018mentornet,wei2020combating}.

In light of this, our main contribution is \textbf{Alternate Experience Replay} (\textbf{AER}), a novel CL optimization scheme that alternates steps of \textit{buffer learning} and \textit{buffer forgetting} to encourage the separation of clean and noisy samples in the buffer. To the best of our knowledge, our work is the first one that exploits forgetting to purify the memory buffer while learning from an online stream. Furthermore, to take advantage of the enhanced separation brought by AER, we propose \textbf{Asymmetric Balanced Sampling} (\textbf{ABS}): a new sample selection strategy designed to select mostly clean samples while keeping the most informative ones from the past. Through extensive experiments, we \textit{i)} evaluate our method across various noise types (both synthetic and real) and rates, \textit{ii)} compare with existing CLN methods adapted to our setting, \textit{iii)} prove the applicability of AER and ABS on other rehearsal-based methods, \textit{\textbf{iv)}} validate the impact of each component on performance and buffer quality.
\section{Related works}
Continual Learning methods can be broadly categorized into regularization-based -- these limit changes to key task-related parameters~\cite{kirkpatrick2017overcoming,zenke2017continual} -- and rehearsal-based methods~\cite{van2019three}.

\noindent{\textbf{Rehearsal}}. In most existing CL scenarios~\cite{chaudhry2019tiny,van2019three,buzzega2020dark}, it has been shown that supplementing current training data with past samples is more effective at mitigating forgetting than any regularization-based methods. A simple yet effective method is Experience Replay (ER)~\cite{ratcliff1990connectionist,robins1995catastrophic} which interleaves the current training batch with past examples. Otherwise, GDumb~\cite{prabhu2020gdumb} pushes this concept to the extreme by greedily storing incoming samples and subsequently training a model from scratch using only the samples stored in the buffer.

\noindent{\textbf{Sampling strategies}}.~Given their low capacity, buffers need to contain a balanced outlook of all seen classes. For this purpose, many employ \textit{reservoir sampling}~\cite{vitter1985random} to update the memory~\cite{buzzega2020dark,bang2021rainbow,caccia2022new}. The outcome is an independent and identically distributed snapshot of the incoming tasks. However, not every sample comes with the same significance or robustness against forgetting. As highlighted by~\cite{buzzega2020rethinking,bang2021rainbow,aljundi2019gradient}, retaining complex samples is crucial for preserving the performance, which they detect through their loss value or model uncertainty, respectively.
\subsection{Learning with Noisy Labels}
Noisy data can originate from multiple sources, including systematic or measurement errors encountered when retrieving historical data~\cite{karimi2020deep,porrello2019spotting,bonicelli2023spotting}, errors introduced by human annotators~\cite{tanno2019learning,hattab2023scoring}, or the presence of outliers~\cite{brodley1999identifying}. Furthermore, noisy labels pose a significant challenge in medical imaging~\cite{karimi2020deep}, where small and noisy validation sets can hinder the effectiveness of model calibration techniques~\cite{ding2021local,2021ICPR_kidney}. Various approaches have been proposed to mitigate the impact of data noise, including adversarial training, regularization and robust loss functions~\cite{hendrycks2018deep,liu2020early,zhang2018generalized}. Additionally, ensemble methods~\cite{nguyen2020self,lu2022selc,2021IETCV} have been proven effective in reducing the impact of noisy data on model performance. 

\tit{Noisy label detection.} The prevalent approach for identifying noisy data is grounded on the memorization effect~\cite{arpit2017closer,jiang2018mentornet}, according to which correctly labelled (\textit{clean}) instances tend to produce a smaller loss than mislabeled (\textit{noisy}) ones during the initial stages of training. However, as the training ensues and the model starts to learn wrong patterns from the noisy data, its predictions become less reliable (\textit{confirmation bias}). In this regard,~\cite{han2020sigua} performs gradient ascent on the noisy samples, building on top of existing sample selection strategies and enhancing loss correction algorithms. Other works exploit separate models to perform sample selection, training either on a probably clean subset (CoTeaching~\cite{han2018coteaching}, MentorNet~\cite{jiang2018mentornet}) or on all seen samples with semi-supervised objectives (DivideMix~\cite{li2020dividemix}, ~\cite{arazo2019unsupervised}).
\subsection{Continual Learning under Noisy Labels}
Recent studies~\cite{bang2022online,kim2021continual,karim2022cnll} conducted in the online CL setup have shown that existing sampling strategies fail to produce meaningful gains in noisy scenarios.
In this respect, the authors of PuriDivER~\cite{bang2022online} propose a sampling strategy that promotes a trade-off between \textit{purity} and \textit{diversity} for samples in the buffer. Methods like SPR~\cite{kim2021continual} and CNLL~\cite{karim2022cnll} use multiple buffers to gradually isolate clean samples: an auxiliary -- usually larger -- buffer gathers data from the stream, while a refinement procedure based on the small-loss criterion extracts only the clean samples into a purified buffer. Afterwards, SPR trains a network using a self-supervised loss on samples from both buffers, while CNLL adopts a semi-supervised approach inspired by FixMatch~\cite{sohn2020fixmatch}. These models are limited to online settings due to either high computational demands or the tendency to overfit, losing effectiveness.
\section{Method}
\label{sec:method}
\tit{Problem setting.} We define the Continual Learning framework as the process of learning from a sequential series of $T$ tasks. During each task $t \in \{0,1, \dots, T-1\}$, input samples $\mathbf{X_t}$ and their annotations $\mathbf{Y_t}$ are drawn from an i.i.d.\ distribution $\mathcal{D}_t$. We follow the well-established class-incremental scenario~\cite{van2019three,farquhar2018towards,buzzega2020dark} in which $\mathbf{{Y}}_{t-1}\cap\mathbf{{Y}}_{t}=\emptyset$ and at task $t$ the learner $f_\theta$ is required to distinguish between all observed classes. Ideally, we wish to minimize:
\begin{equation}
    \theta^*=\argmin_{\theta}{\mathbb{E}_t{\bigg{[}\operatornamewithlimits{\mathbb{E}}_{\mathcal{B}\sim\mathcal{D}_t}{\Big{[}\mathcal{L}(f_\theta(\mathbf{x}),y)\Big{]}}\bigg{]}}},
\end{equation}
where $\mathcal{L}$ is the cross-entropy loss and $\mathcal{B}=(\mathbf{x},y)$. As in CL the objective above is inaccessible, we leverage a fixed-size buffer $\mathcal{M}$ to store and replay part of the incoming samples. As a result, the generalized objective for rehearsal CL can be defined as:
\begin{equation}
    \label{equ:rehersal}
    \theta^*=\argmin_{\theta}{\operatornamewithlimits{\mathbb{E}}_{\mathcal{B}\sim\mathcal{D}_t}{\Big{[}\mathcal{L}(f_\theta(\mathbf{x}),y)\Big{]}}} + \mathcal{L}_R,
\end{equation}
where the \textit{replay regularization} term $\mathcal{L}_R$ depends on the choice of the replay-based method. Although our approach can be equally applied to advanced choices of $\mathcal{L}_R$~\cite{buzzega2020dark,bonicelli2022effectiveness,hou2019learning,boschini2022transfer} (see~\cref{sec:ablation}), in this work we build upon the simplest strategy and leverage Experience Replay~\cite{ratcliff1990connectionist,robins1995catastrophic}:
\begin{equation}
    \label{equ:experience_replay}
\mathcal{L}_R=\operatornamewithlimits{\mathbb{E}}_{(\mathbf{x}_r,y_r)\sim\mathcal{M}}{\Big{[}\mathcal{L}(f_\theta(\mathbf{x}_r),y_r)\Big{]}}.
\end{equation}
As the objective in~\cref{equ:experience_replay} could result in bias accumulation toward the current task~\cite{ahn2020ssilss}, we adopt the asymmetric cross-entropy loss introduced in~\cite{caccia2022new}.
\subsection{Alternate Experience Replay (AER)}
\label{sec:aer}
\begin{algorithm}[t]
\caption{Overall procedure of AER with ABS}
    \label{alg:pseudocode}
    \textbf{Input}: stream data $\mathcal{D}_t$, buffer $\mathcal{M}$, training epochs $T$
    \begin{algorithmic}[1]
    \For {epoch in $1, \ldots, T$}
\State $\theta_{\operatorname{CHK}} \gets \theta_{\operatorname{epoch}}$ \Comment{save current parameters of $f_{\theta}(\cdot)$}
\For {batch $\mathcal{B} \sim\mathcal{D}_t$}
\State $p \gets \text{normalize}(\{s(x); \forall (x,\tilde{y}) \in \mathcal{M}\})$ \Comment{compute asymmetric scores (\textit{selection})}
\If {epoch in $T_{\text{on}}$} 
\State train on $\mathcal{B} \cup (\mathbf{x}, \tilde{y})\sim\mathcal{M}$ \Comment{buffer learning}
\Else 
\State train on $\mathcal{B}\sim\mathcal{D}_t$ \Comment{buffer forgetting}
\State $\mathcal{R}\gets \text{reservoir}(\{(\mathbf{x}, \tilde{y})\in\mathcal{B}: \mathcal{L}(\mathbf{x}, \tilde{y})<\mathcal{L}_{\alpha}\})$ \Comment{sample insertion}
\State $\mathcal{M}[z \sim p]\gets \mathcal{R}$ \Comment{replace data sampled with $p$ with stream data $\mathcal{R}$}
\EndIf
\EndFor
\If {epoch in $T_{\text{off}}$} 
\State $\theta_{\operatorname{epoch}} \gets \theta_{\operatorname{CHK}}$ \Comment{restore previous model checkpoint}
\EndIf
\EndFor

    \end{algorithmic}
\end{algorithm}
As discussed above, we tackle the challenge of \textbf{Continual Learning under Noisy Labels} (\textbf{CLN}), where each incoming dataset is affected by noise in the labeling process, leading to mislabeled training examples. For a given instance $\mathbf{x}_i\in\mathcal{D}_t$, we indicate with $\tilde{y}_i\sim{\tilde{Y}_i}$ the labels corrupted with annotation noise and with $\operatorname{Pr}{(\tilde{y}_i\neq y_i)}$ the respective \textit{noise rate}. In this setting, we must simultaneously address the challenges posed by both noisy labels $\mathbf{\tilde{Y}}_t$ and the problem of forgetting. To achieve this, our main focus is on constructing a memory set $\mathcal{M}$ that is as clean and representative as possible. Since this objective involves distinguishing between noisy and clean examples when populating the memory set, our methodology seeks to maintain a significant gap between the losses of clean and noisy samples. This gap is indeed crucial for filtering examples through the widely used small-loss criterion. However, with no countermeasure, the loss gap starts to deteriorate as the replay of a small selection of data ensues (see~\cref{fig:buffer_loss_divergence}, left). This effect is exacerbated in the popular offline (\ie multi-epoch) CL setting~\cite{rebuffi2017icarl,wu2019large,buzzega2020dark}, where we might be forced to trade-off convergence on the current task to avoid overfitting the mislabeled samples~\cite{zhang2016understanding,arpit2017closer}.

To counteract the vanishing effect of the small-loss criterion and encourage the separation between the losses of noisy and clean samples, our novel methodology named \textbf{Alternate Experience Replay} (\textbf{AER}) induces forgetting of buffer datapoints. We refer the reader to~\cref{alg:pseudocode} for a summary of the overall procedure. Specifically, we divide the training epochs for the current task into two categories: \textbf{buffer learning} and \textbf{buffer forgetting} epochs. The training process involves alternating between these two modes of learning.

\begin{itemize}[noitemsep]
    \item \textbf{Buffer learning.}~In this regime, we train the model with standard replay (line 6) as in~\cref{equ:rehersal}. Importantly, we do not modify the samples stored in the memory buffer $\mathcal{M}$ (no insertion or removal operations are performed).
    \item \textbf{Buffer forgetting.}~In this case (line 8), we omit regularization on the memory buffer and focus the training exclusively on data from $\mathcal{D}_t$. By halting regularization and causing the subsequent forgetting of buffer datapoints, the loss of noisy examples is likely to increase more rapidly than that of clean ones~\cite{toneva2018an,maini2022characterizing}. This, in turn, makes the small-loss criterion reliable once again (see~\cref{fig:buffer_loss_divergence}, right). On top of that, we update $\mathcal{M}$ (line 9) through a loss-based selection strategy during these epochs (see \cref{sec:sample_selection}).
\end{itemize}

This way, at the end of each buffer forgetting epoch, we get a cleaner version of the memory buffer. However, cycling between buffer learning and forgetting could result in the buffer being under-optimized, as it is effectively exploited only during the former epochs. We avoid this through \textit{model checkpointing}: specifically, at the start of each forgetting epoch, we save the parameters of the model $f_\theta$ (line 12) and restore them at the end of the same epoch (line 2). While this option results in the model being optimized for only half of the epochs, we prove in~\cref{sec:experiments} that the trade-off significantly enhances the final accuracy of the model.
\subsection{Asymmetric Balanced Sampling (ABS)}
\label{sec:sample_selection}
\begin{figure}[t]
\centering
\includegraphics[width=0.9\textwidth]{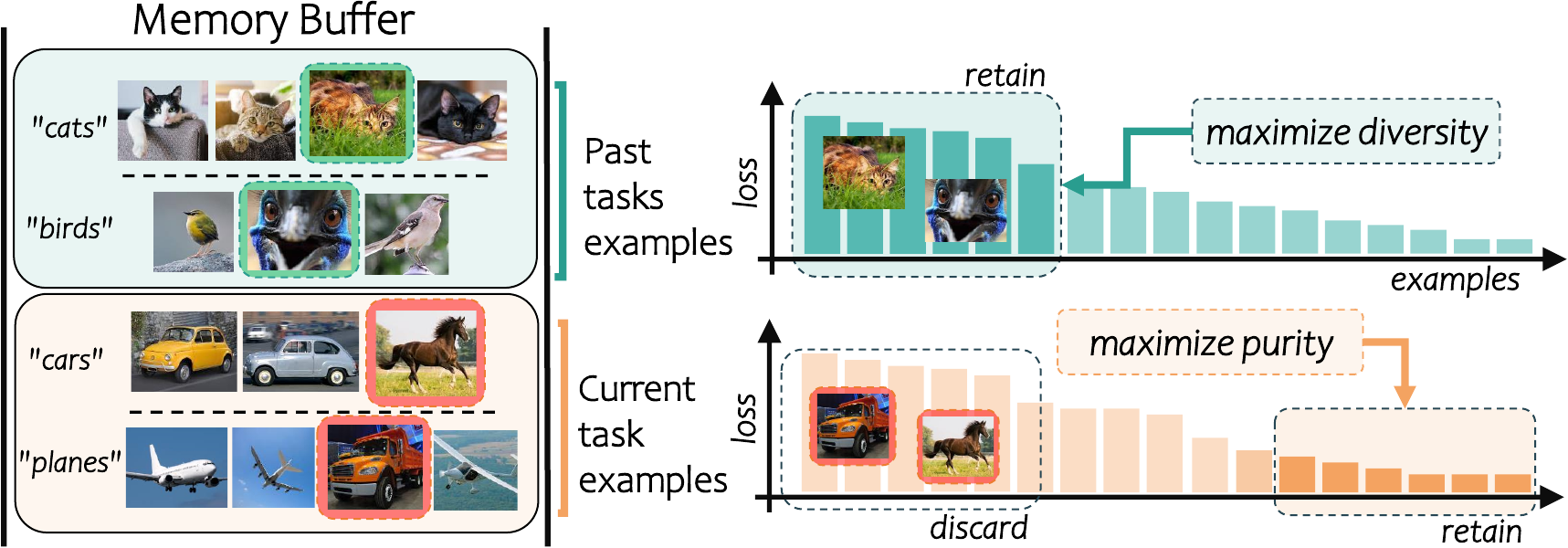}
\caption{Asymmetric Balanced Sampling (ABS). Past examples are chosen to retain the most complex ones, while the criterion is reversed for the current task to maximize purity.}
\label{fig:alternate_replay}
\end{figure}
In this section, we outline the sampling strategy used to \textit{insert} and \textit{delete} examples into and from the memory buffer during each buffer forgetting epoch.

\tit{Sample insertion.}~Given a batch of data $\mathcal{B}$ from the current task, the first step is to determine which examples should be included in the buffer. To encourage the inclusion of clean examples, we exploit the memorization effect and employ a simple criterion that involves applying a threshold to the loss value.
Formally, let $\alpha$ denote the percentage of samples within the current batch that we intend to exclude from the insertion procedure, we compute:
\begin{equation}
\mathcal{R}=\{(\mathbf{x}, \tilde{y})\in\mathcal{B}: \mathcal{L}(\mathbf{x}, \tilde{y})< \mathcal{L}_{\alpha}\}
\end{equation}
where $\mathcal{L}_{\alpha}$ is the loss value at the $\alpha$-th percentile of the loss distribution over $\mathcal{B}$. For our experiments, we set $\alpha$ to $75$, thus discarding the $75\%$ of samples with the highest loss and treating the remaining $25\%$ as candidates to be inserted in the buffer (lines 9-10 of~\cref{alg:pseudocode}).

\tit{Sample selection.}~We approach the selection process by sampling from a probability distribution $p(\mathbf{x})$ defined over all exemplars $\forall\mathbf{x}\in\mathcal{M}$ in the buffer. To model such a distribution, as carried out by most methods~\cite{buzzega2020rethinking,bang2021rainbow,bang2022online}, we leverage the score $s(x) = \mathcal{L}(x, y) \geq 0$ given by the loss function. It is noted that a valid distribution can be then obtained by normalizing these scores, such that $p(x)=\frac{s(x)}{Z}$ where $Z=\sum_{x\in\mathcal{M}}{s(x)}$. We refer to this strategy as \textbf{Loss-Aware Symmetric Sampling} (\textbf{LASS}). In LASS, examples with higher loss are more likely to be replaced, leading to a memory buffer that maintains greater \textbf{purity}.

However, we argue that a replacement criterion based on loss value like LASS could be detrimental in terms of \textbf{diversity}, as it discourages the retention of complex yet clean samples into the memory buffer. Indeed, these examples tend to exhibit higher loss value, as they lie in the proximity of the decision boundary~\cite{aljundi2019gradient,buzzega2020rethinking,bang2021rainbow}. Conversely, we argue that the small-loss criterion should be slightly revised when employed for memory replacement. Indeed, an ideal replacement strategy should preserve challenging yet representative examples from past tasks, in order to ensure an effective regularization signal.

In light of this, we propose a novel replacement strategy called \textbf{Asymmetric Balanced Sampling} (\textbf{ABS}, see~\cref{fig:alternate_replay}) that looks for a compromise between two contrasting objectives. Namely, it aims to ensure the inclusion of both \textit{ii)} high-loss (i.e., complex) samples from the past and \textit{i)} small-loss (i.e., clean) samples from the present. To do so, it builds upon an \textbf{asymmetric score} (line 4) that differentiates whether a given example in the memory buffer belongs to a past task or the current one. Specifically, the score $s(x)$ is equal to the loss $\mathcal{L}(x, y)$ if the example comes from the current task $\mathcal{M}_{cur}$ (as in LASS). Conversely, to encourage diversity, the criterion is \textbf{reversed} for examples from past tasks $\mathcal{M}_{past}$, with the score being equal to $-\mathcal{L}(x, y)$. By taking this approach, we accept that a few mislabeled examples from past tasks might remain in the memory buffer; however, given the joint effect of the insertion policy and the LASS-like side of the replacement criterion -- both of which tend to favor purity among examples from the current task -- we can be cautiously optimistic that the examples from past tasks are correctly annotated (see~\cref{sec:ablation} for an empirical analysis).

Finally, to achieve a balanced representation of both current and previous tasks in the memory buffer, we decide whether to replace a sample from the current or previous task based on their relative sizes. Considering $|\mathcal{M}_{cur}|$ as the number of samples from the current task in $\mathcal{M}$, we define a Bernoulli distribution with probability $\frac{|\mathcal{M}_{cur}|}{|\mathcal{M}|}$, where a success corresponds to sampling from $\mathcal{M}_{cur}$. This approach ensures that the likelihood of replacing a sample is proportional to the current task's sample size within the entire memory buffer.
\subsection{Buffer consolidation}
While combining AER and ABS achieves a balance between sample purity and complexity preservation, a further reduction in noise levels can be achieved by optimally selecting samples from $\mathcal{M}$ at the end of each task. We employ a MixMatch-based~\cite{berthelot2019mixmatch} approach to enhance model robustness, utilizing uncertain samples as unlabeled ones (\ie buffer \textit{consolidation} in the following). Further details about this phase are provided in the supplementary materials.
\section{Experiments}
\label{sec:experiments}
\tit{Datasets and noise settings.}~We conduct experiments on five distinct datasets and various levels of noise. Specifically, we use the \textbf{Seq. CIFAR-100} dataset~\cite{krizhevsky2009learning}, which contains $32\times32$ images from $100$ categories, split into $10$ tasks, and the \textbf{Seq. NTU RGB+D}~\cite{shahroudy2016ntu} dataset for 3D skeleton-based human action recognition, featuring $60$ classes divided into $6$ tasks. On these datasets, we inject two types of synthetic noise commonly employed in literature~\cite{li2020dividemix,jiang2018mentornet,han2018coteaching}: \textit{symmetric} and \textit{asymmetric} noise. In the first scenario, we replace the ground-truth label with probability $r \in [0, 1]$ determined by the designated noise rate. The asymmetric or class-dependent noise setting, instead, is an approximation of real-world corruption patterns, altering labels within the same superclass as in~\cite{patrini2017making,zhang2018generalized}. To further address real-world label noise, we evaluate our method on \textbf{Seq.\ Food-101N}~\cite{lee2018cleannet} ($5$ tasks), composed of images gathered from the web, thus containing \textit{instance-level} annotation noise. Additionally, ResNet18~\cite{he2016deep} is used for Seq.\ CIFAR-100 with 50 epochs per task, ResNet34~\cite{he2016deep} for Food-101N with 20 epochs, and EfficientGCN-B0~\cite{song2021constructing} for Seq.\ NTU-60 with 30 epochs.

\tit{Benchmarking.}~In line with notable CL works~\cite{rebuffi2017icarl,hou2019learning,wu2019large,bang2021rainbow,boschini2022class,menabue2024semantic,frascaroli2024clip}, we adhere to a class-incremental and \textbf{multi-epoch} setting, in which samples can be experienced multiple times within the respective task. The results are presented in terms of Final Average Accuracy (FAA), computed at the end of the last task. All results are averaged across 5 runs.

We compare against PuriDivER~\cite{bang2022online}, the current state-of-the-art selection strategy for CLN, as well as common rehearsal CL baselines. For the latter, we follow~\cite{bang2022online} and apply both CoTeaching~\cite{han2018coteaching} and DivideMix~\cite{li2020dividemix} to consolidate the buffer of ER~\cite{robins1995catastrophic,ratcliff1990connectionist} and GDumb~\cite{prabhu2020gdumb}. Since current CLN methods are designed for the online setting (\ie a single training epoch is allowed), a direct comparison would be problematic: based on~\cref{sec:aer}, we hence refine PuriDivER by suspending memory updates after the first epoch, naming such method as \textbf{PuriDivER.ME}. We also compare with SPR~\cite{kim2021continual} and CNLL~\cite{karim2022cnll}, adapted for offline CLN and with the same overall memory budget for fairness. Given the huge computational demands of SPR and CNLL, evaluating them on complex datasets like CIFAR-100 and NTU proved impractical: hence, we employ the smaller Seq. CIFAR-10 dataset (5 tasks).

Finally, the upper bound is attained by training jointly on all tasks (\textit{Joint}), while the lower bound is attained by training without any countermeasure to forgetting or noise (\textit{Finetune}).
\begin{table*}[t!]
\begin{center}    
\renewcommand{\fapp}[2]{\ensuremath{#1}\tiny{$\pm$ \ensuremath{#2}}}    
\centering
\setlength\tabcolsep{2pt}
\resizebox{\linewidth}{!}{
\begin{tabular}{lccc|cc|cc}
\hline
\noalign{\smallskip}
\textbf{Benchmark} & \multicolumn{5}{c}{\textbf{Seq.\ CIFAR-100}} & \multicolumn{2}{c}{\textbf{Seq.\ NTU-60}} \\
\hline

&\multicolumn{3}{c}{\textit{symm}} &\multicolumn{2}{c}{\textit{asymm}} &\multicolumn{2}{c}{\textit{symm}}\\

\textbf{Noise rate} & 20 & 40 & 60 & 20 & 40 & 20 & 40 \\
\hline
\noalign{\smallskip}
Joint       & \fapp{54.77}{0.61} & \fapp{38.46}{0.92} & \fapp{23.36}{1.09} & \fapp{56.70}{0.57} & \fapp{42.61}{0.92} & \fapp{68.26}{0.69} & \fapp{63.02}{0.88}  \\
Finetune    & \fapp{08.65}{0.13} & \fapp{07.55}{0.14} & \fapp{06.15}{0.17} & \fapp{07.78}{0.14} & \fapp{05.73}{0.09} & \fapp{14.30}{0.51} & \fapp{11.73}{1.07}   \\
\hline
\noalign{\smallskip}
ER~\cite{vitter1985random}                            & \fapp{25.14}{0.28} & \fapp{14.64}{0.23} & \fapp{8.92}{0.23} & \fapp{29.42}{0.39} & \fapp{18.91}{0.86}   & \fapp{29.95}{2.16} & \fapp{16.02}{0.27} \\
\hspace{.3em} + \textit{CoTeaching}~\cite{han2018coteaching} & \fapp{25.79}{0.61} & \fapp{14.46}{0.49} & \fapp{8.92}{0.30} & \fapp{32.18}{2.55} & \fapp{20.76}{2.44}   & \fapp{43.87}{0.78} & \fapp{30.71}{1.86} \\
\hspace{.3em} + \textit{DivideMix}~\cite{li2020dividemix}    & \fapp{33.31}{0.27} & \fapp{22.91}{0.43} & \fapp{13.58}{1.0.2} & \fapp{36.98}{0.78} & \fapp{26.10}{1.10} & \fapp{40.92}{0.97} & \fapp{32.07}{1.73} \\
\hline
\noalign{\smallskip}
GDumb~\cite{prabhu2020gdumb}                                 & \fapp{16.96}{0.61} & \fapp{11.31}{0.45} & \fapp{7.62}{0.28} & \fapp{17.25}{0.28} & \fapp{11.75}{0.06}  & \fapp{11.34}{0.21} & \fapp{6.86}{0.86} \\
\hspace{.3em} + \textit{CoTeaching}~\cite{han2018coteaching} & \fapp{17.02}{0.50} & \fapp{13.17}{0.31} & \fapp{8.17}{0.99} & \fapp{17.07}{0.54} & \fapp{12.05}{0.62}  & \fapp{12.37}{2.04} & \fapp{8.82}{0.51} \\
\hspace{.3em} + \textit{DivideMix}~\cite{li2020dividemix}    & \fapp{19.26}{0.97} & \fapp{15.67}{0.97} & \fapp{10.51}{0.32} & \fapp{18.80}{1.55} & \fapp{13.29}{0.29} & \fapp{15.96}{1.16} & \fapp{7.49}{1.11} \\
\hline
\noalign{\smallskip}
PuriDivER~\cite{bang2022online}  & \fapp{27.53}{0.53} & \fapp{24.36}{0.40} & \fapp{17.81}{0.43} & \fapp{25.46}{1.44} & \fapp{18.84}{0.64}             & \fapp{39.33}{1.59} & \fapp{38.86}{0.79} \\ 
PuriDivER.ME\textsuperscript{\textbf{\dag}}  & \fapp{41.25}{0.63} & \fapp{37.61}{0.85} & \faau{27.18}{0.76} & \faau{41.65}{0.49} & \faau{30.22}{0.74} & \fapp{43.10}{1.11} & \fapp{38.07}{1.06} \\ 
\hline
\noalign{\smallskip}
\textbf{OURs}               & \faau{44.34}{0.48} & \faau{38.64}{0.57} & \fapp{26.34}{0.85} & \fapp{41.24}{0.40} & \fapp{29.26}{0.91}                        & \faau{47.71}{0.89} & \faau{43.11}{2.12}  \\
\hspace{.3em} \textit{w. consolidation} & \fappb{46.11}{1.46}  &  \fappb{40.27}{0.40}  &  \fappb{34.81}{1.63}  &  \fappb{43.67}{0.73}  &  \fappb{32.64}{0.48} & \fappb{48.73}{1.20} & \fappb{45.19}{0.05} \\
\hline
\end{tabular}
}
\end{center}
\caption{Final Average Accuracy (FAA) $[\uparrow]$ on multiple datasets and noise rates. \textsuperscript{\textbf{\dag}} Additional baselines adapted to the multi-epoch scenario.}
\label{table:ss_faa_new}
\end{table*}
\subsection{Comparison with State-of-the-Art}
\label{sec:CNL_compare}
The results of our main evaluation are presented in~\cref{table:ss_faa_new}. To streamline the discussion, we first compare our approach with traditional continual learning baselines, followed by an analysis of methods designed for continual learning under noisy labels (\eg PuriDivER).

\tit{Comparison with rehearsal baselines.}~As outlined by~\cref{table:ss_faa_new}, the approaches relying solely on buffer consolidation -- such as ER and GDumb -- are poorly effective, especially as noise levels rise. Regarding GDumb, its training phase is limited to the content of the memory buffer, preventing it from utilizing the data variety available throughout the task. This limitation is also evident from the comparison with standard ER, which consistently outperforms GDumb when noise levels are low. These outcomes highlight the benefits of performing multiple training iterations. However, this advantage turns into a double-edged sword as the stream becomes noisier, leading to a significant drop in performance.
\begin{table}[t]
\centering    
    \begin{minipage}[b]{0.48\textwidth}
    \centering
    \begin{center}
    \setlength\tabcolsep{3pt}
\begin{tabular}{lccc}
\hline\noalign{\smallskip}
\multicolumn{4}{c}{\textbf{Seq.\ CIFAR-10 -- 40\% \textit{symm}}} \\
\hline
\noalign{\smallskip}
\multicolumn{2}{c}{\textbf{Buffer size (total)}} & 2500 & \textit{unlimited} \\
\hline
\noalign{\smallskip}
CNLL & \textit{1 epoch} & $38.14$ & $57.26$ \\
CNLL & \textit{50 epochs} & $35.46$ & $43.43$ \\
\textbf{OURs} & \textit{50 epochs} & $\mathbf{67.10}$ & $\mathbf{76.83}$ \\
\hline
\noalign{\smallskip}
\multicolumn{2}{c}{\textbf{Buffer size (total)}} & \multicolumn{2}{c}{1000}\\
\hline
\noalign{\smallskip}
SPR\textsuperscript{\textdaggerdbl} & \textit{25 epochs} & \multicolumn{2}{c}{$26.34$}  \\
\textbf{OURs} & \textit{25 epochs} & \multicolumn{2}{c}{$\mathbf{63.65}$} \\
\hline
\end{tabular}
    \end{center}
    \captionof{table}{Comparison with SPR and CNLL. \textsuperscript{\textdaggerdbl}training iterations spread across epochs.}
    \label{tab:sprcnll}
    \end{minipage}
    \hfill
\begin{minipage}[b]{0.48\textwidth}
    \centering
    \begin{center}
    \setlength\tabcolsep{3.5pt}
\begin{tabular}{ccccc|c}
\hline\noalign{\smallskip}
\multicolumn{6}{c}{\textbf{Seq.\ CIFAR-100 -- 60\% \textit{symm}}} \\
\hline
\noalign{\smallskip}
ER & w. ACE & $\alpha$ & AER & ABS & \textbf{FAA} \\
\hline
\noalign{\smallskip}
\cmark & \cmark &   &   &   & 11.65 \\
\cmark & \cmark & \cmark &   &   & 19.97 \\
\cmark & \cmark  & \cmark & \cmark &   & 24.19 \\
\cmark & \cmark & \cmark &  & \cmark & 21.68\\
\cmark & \cmark & \cmark & \cmark & \cmark & \textbf{26.34}\\
\cmark &  & \cmark  & \cmark & \cmark & 22.02 \\
\hline
\end{tabular}

    \end{center}
    \captionof{table}{Ablation study for each component of our proposal -- 60 \% symmetric noise.}
    \label{table:model_ablation}
\end{minipage}
\end{table}

\tit{Comparison with CNL methods.}~Firstly, we highlight the substantial improvement achieved by our adapted PuriDivER.ME, which outperforms PuriDivER by an average of $8.36\%$. Both versions perform buffer consolidation~\cite{bang2022online} at the end of each CL task; however, PuriDivER relies on a model trained over multiple epochs, which leads to the degradation of the small-loss criterion, an issue outlined in~\cref{sec:aer}. Moreover, both PuriDivER.ME and ER + DivideMix are consistently surpassed by our proposal. In particular, we measure an average $1.50\%$ gain over the best competitor's performance without any buffer consolidation, suggesting that our proposal improves the purity and diversity of samples in the buffer. However, as the sample selection is not perfect, applying an additional buffer consolidation technique tends to be more effective in more complex noise scenarios, with an average improvement of $4.71\%$. 

We conduct additional comparisons with CNLL and SPR (\cref{tab:sprcnll}) and PuriDivER.ME (\cref{table:food}). For the latter, we adopt the more realistic Food-101N dataset (\ienospace, images collected from the web and automatically labeled). Even in these scenarios, our approach remains superior, both with and without buffer consolidation. We remark that these considerable gains come with a remarkable speed-up in terms of both time and resources used (see supplementary materials), making it more suitable for a multi-epoch incremental scenario.

\begin{table}[t]
\begin{minipage}{\textwidth}
    \vspace{-1em}
    \begin{minipage}[b]{0.50\textwidth}
    \begin{center}
    \renewcommand{\fapp}[2]{\ensuremath{#1}$\pm$\ensuremath{#2}}
\setlength\tabcolsep{5pt}
\setlength\tabcolsep{5pt}
\begin{tabular}{lc}
\hline\noalign{\smallskip}
\textbf{Benchmark} & \textbf{Food-101N} \\
\hline
\noalign{\smallskip}
Joint & \fapp{39.91}{1.05} \\
\hline
\noalign{\smallskip}
\textbf{PuriDivER.ME} & \fapp{28.62}{0.85}\\
\hline
\noalign{\smallskip}
\textbf{OURs} & \fapp{29.86}{1.18} \\
\hspace{.3em} \textit{w.\ buffer fit.} & \fappb{34.79}{0.64} \\
\hline
\end{tabular}
    \end{center}
    \vspace{-0.5em}
    \captionof{table}{Performances $[\uparrow]$ of our method and main competitor on a real-world noisy dataset.}
    \label{table:food}
    \centering
    \includegraphics[width=\textwidth]{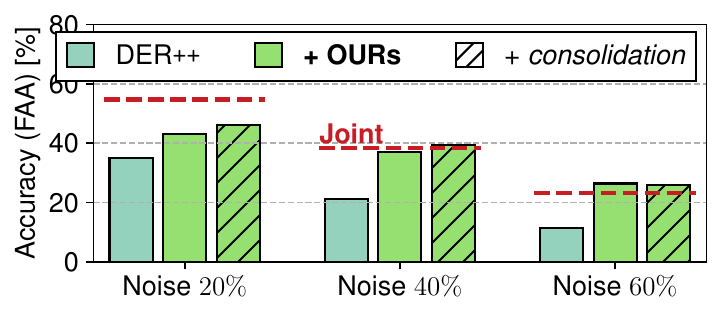}
    \vspace{-2.5em}
    \captionof{figure}{FAA ($[\uparrow]$) of \dpp with our method and buffer fitting.}
    \label{fig:derpp_abl}
    \end{minipage}
    \hfill
\begin{minipage}[b]{0.48\textwidth}
    \vspace{1.65em}
    \centering
    \includegraphics[width=\textwidth]{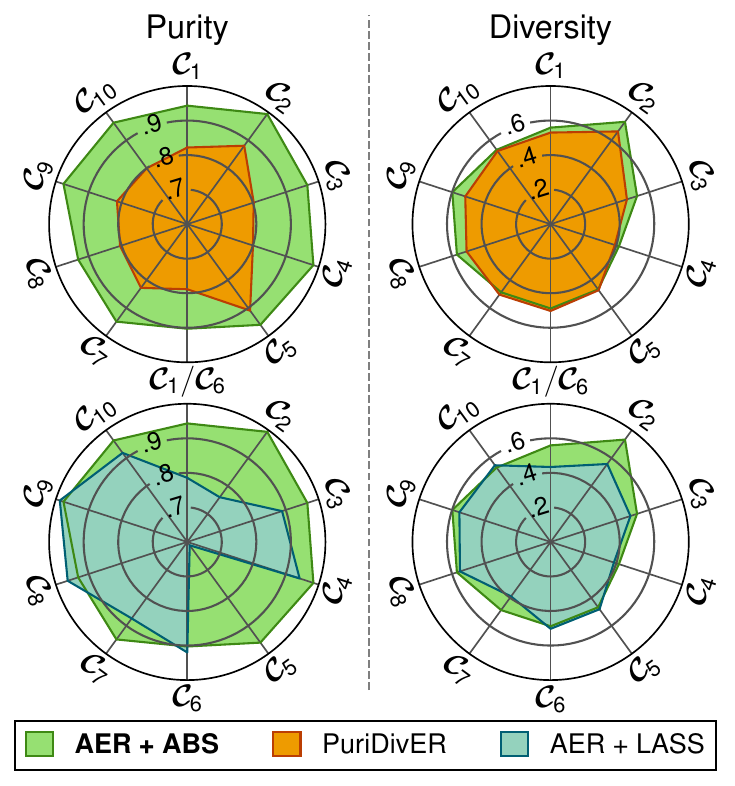} 
    \vspace{-2em}
    \captionof{figure}{Final composition of the buffer with different choices of sample selection.}
    \label{fig:purity_diversity}
\end{minipage}
\end{minipage}
\end{table}
\section{Model analysis}
\label{sec:ablation}
\tit{Ablative study.}~We herein aim to investigate the impact of each component. Starting from the base rehearsal method used in our research, \ie ER-ACE~\cite{caccia2022new}, we gradually introduce our two main contributions, AER and ABS, one at a time. As seen from the results in~\cref{table:model_ablation}, each additional feature produces an increase in performance on Seq. CIFAR-100. For an in-depth analysis of the effects of the asymmetric cross-entropy loss function (ACE), we compare against the standard cross-entropy (\ie ER in~\cref{table:model_ablation}). The results indicate that the contribution of ACE is significant, aligning with both~\cite{caccia2022new} and our initial expectations.

\tit{Purity of the buffer.}~Considering Seq.\ CIFAR-10 ($40\%$ noise),~\cref{fig:purity_diversity} depicts the \textit{purity} and the \textit{diversity} of the buffer produced by ABS, PuriDivER.ME, and LASS. For each class, purity is defined as the ratio of examples labeled correctly within the memory buffer. Instead, we model diversity as the intra-class variation within each class, thereby computing the average std. deviation of the features produced by the \textit{Joint} ideal model. Finally, we scale all metrics according to their occurrence rate to account for potential imbalances in the number of examples from different classes. As shown in~\cref{fig:purity_diversity}, ABS clearly outperforms both LASS and PuriDivER.ME in terms of purity and diversity. Unexpectedly, LASS yields a particularly unbalanced buffer, with only the most recent classes showing a good balance. In contrast, PuriDivER.ME achieves better balance but falls short in terms of purity.

\tit{Applicability to other methods.}~To evaluate whether AER/ABS can enhance other rehearsal methods, we apply them on \textbf{DER++}~\cite{buzzega2020dark} and conduct tests on Seq. CIFAR-100. We also report the results with and without the consolidation phase (\cref{sec:CNL_compare}). The gains shown in~\cref{fig:derpp_abl} support the validity of our AER/ABS on enhancing other CL baselines.

\tit{Additional results.} In the supplementary materials, we provide: \textit{i)} details about the experimental settings, the adapted baselines, noise injection process and hyperparameters, \textit{ii)} the evaluation of Final Forgetting (FF), \textit{iii)} an analysis of the computational costs, \textit{iv)} an evaluation of the speed at which the model learns the noisy data, \textit{v)} a sensitivity analysis conducted on the hyperparameter $\alpha$, which controls the purity within the sample insertion strategy.

\section{Conclusions}
We present an innovative framework for Continual Learning in the presence of Noisy Labels, a common issue in real-world AI applications. We focus on the multi-epoch class-incremental scenario, arguing the shortcomings of current methods leveraging the small-loss criterion. We hence appeal to a long-standing enemy of continual learning -- \textit{forgetting} -- and propose Alternate Experience Replay to maintain a clear separation between mislabeled and clean samples. Additionally, we introduce Asymmetric Balanced Sampling to enhance sample diversity and purity within the buffer. We demonstrate the merits of our approach through extensive experiments, showcasing its potential in noisy incremental scenarios.
\section*{Acknowledgements}
This paper has been supported from Italian Ministerial grant PRIN 2020 “LEGO.AI: LEarning the Geometry of knOwledge in AI systems”, n. 2020TA3K9N. We acknowledge ISCRA for awarding this project access to the LEONARDO supercomputer, owned by the EuroHPC Joint Undertaking, hosted by CINECA (Italy). Finally, the authors would like to express their sincere gratitude to Alberto Zurli for his earlier work on this topic and his valuable, constructive contributions to the discussions.


\bibliography{egbib}
\newpage

\setcounter{table}{0}
\setcounter{figure}{0}

\renewcommand{\thesubsection}{\thesection.\alph{subsection}}
\renewcommand{\thesection}{\Alph{section}}
\renewcommand{\thetable}{\Alph{table}}%
\renewcommand{\thefigure}{\Alph{figure}}%

\appendix
\section{On the effectiveness of buffer consolidation}
\label{sec:buff_consolidation_details}
By combining AER with ABS we obtain a balance between purity -- for samples of the current task -- while preserving the complexity of those from the past. To achieve this, the backbone network had to be trained on a stream of noisy data. While we find that the effect of noise from the current task is mitigated by AER (\cref{sec:ablation_loss_stream}), we can further reduce its influence with the help of the memory buffer.

In principle, with an ideal sample selection strategy we could simply train on samples from $\mathcal{M}$ to adjust the predictions of the network at the end of the task in a fully-supervised fashion (\textbf{buffer fit.}). While we empirically find in~\cref{sec:experiments} that such a strategy delivers remarkable results, we can refine it to handle more complex noise scenarios.

In particular, we use a modified version of MixMatch~\cite{berthelot2019mixmatch} to obtain a more robust model, using the most \textit{uncertain} samples as a source for unlabeled data. Similarly to~\cite{arazo2019unsupervised}, we fit a two-component Gaussian Mixture Model (GMM) $g(\mathcal{L})$ on the loss $\mathcal{L}$ of each $(\mathbf{x}, \tilde{y})\in\mathcal{M}$. Then, we compute the perceived uncertainty of each sample $u(\mathbf{x})$ as the posterior $g(l|\mathcal{L})$, where $l$ indicates the Gaussian component with the smaller mean.
Samples are then separated into \textit{pure} $\mathcal{P}$ and \textit{uncertain} $\mathcal{U}$ with a simple threshold on $g(l|\mathcal{L})$.

From this, samples in $\mathcal{P}$ have label $\tilde{y}\approx y$, thus we can use them to compute a supervised loss term. Instead, for  $\mathbf{x}\in\mathcal{U}$ we compute $\hat{y}$ using the model's response on different augmentations $T$ of $\mathbf{x}$:
\begin{equation}
\label{equ:mixmatch}
   \hat{y} =  u(\mathbf{x}) \tilde{y} + \frac{1 - u(\mathbf{x})}{\eta}\sum_{i=1}^{\eta}{f_\theta(T(\mathbf{x}))},
\end{equation}

Finally, we obtain the refined set $\mathcal{Q}=\{(\mathbf{x}, \hat{y}): (\mathbf{x}, \tilde{y})\in\mathcal{U}\}$ and follow up with the MixMatch procedure to compute the supervised and self-supervised loss terms $\mathcal{L}_s$ and $\mathcal{L}_u$ respectively. The overall loss term is computed as $\mathcal{L}_s + \lambda_u\mathcal{L}_u$, where $\lambda_u$ is a regularization hyperparameter. 

\section{On the influence of the hyperparameter $\alpha$}

In this section, we want to carry out a sensitivity analysis targeting the value of $\alpha$. Recall that alpha controls the proportion of samples to be discarded from the insertion phase within the buffer. We here report the results yielded by several $\alpha$ values under three different noise settings (asymm. 40\%, symm. 40\%, symm. 60\%). The experiments, reported in~\cref{tab:alpha_influence}, are conducted on Split\ CIFAR-100, with performance measured in terms of Final Average Accuracy. It can be concluded that $\alpha >= 50\%$ is a good choice, with gains that stabilize around $60\% - 70\%$. In our experiments, we remark that we avoided tuning $\alpha$ and set the same value for every dataset/noise ratio/noise type.

\begin{table}[t]
\begin{center}
\setlength{\tabcolsep}{4pt}
    
\begin{tabular}{c|c|c|c|c|c|c}
\hline
\multirow{2}{*}{CIFAR-100} & \multicolumn{6}{c}{\textbf{Parameter} $\mathbf{\alpha}$} \\ \cline{2-7} 
                      & \textbf{0\%}           & \textbf{25\%}           & \textbf{50\%}            & \textbf{60\%}     & \textbf{75\%}               & \textbf{90\%} \\ \hline
\textit{Asym 40\% }            & $24.76$\scriptsize{$\pm$ $0.14$}  & $26.69$\scriptsize{$\pm$ $0.26$} & $28.76$\scriptsize{$\pm$ $0.51$}   &    $29.70$\scriptsize{$\pm$ $0.61$}    &  $29.26$\scriptsize{$\pm$ $0.91$}    & $29.90$\scriptsize{$\pm$ $0.53$}  \\ \hline
\textit{Sym 40\%}    &     $27.01$\scriptsize{$\pm$ $0.84$}   & $31.99$\scriptsize{$\pm$ $0.62$}  & $36.92$\scriptsize{$\pm$ $0.55$}       &  $38.52$\scriptsize{$\pm$ $0.70$}       & $38.64$\scriptsize{$\pm$ $0.57$} &  $39.40$\scriptsize{$\pm$ $0.70$} \\ \hline
\textit{Sym 60\%  }  &  $15.30$\scriptsize{$\pm$ $0.44$}   & $17.35$\scriptsize{$\pm$ $0.06$}    &  $20.74$\scriptsize{$\pm$ $0.48$}     & $24.61$\scriptsize{$\pm$ $0.55$}     &   $26.34$\scriptsize{$\pm$ $0.85$}    & $30.67$\scriptsize{$\pm$ $1.26$} \\ \hline
\end{tabular}
\label{tab:mytable}
\end{center}
\caption{FAA [$\uparrow$] on CIFAR100 with varying noise to assess the influence of $\mathbf{\alpha}$}
\label{tab:alpha_influence}
\end{table}
\section{On the effectiveness of AER as a regularizer for CNL}
\label{sec:ablation_loss_stream}
\begin{figure}[t]
    \centering
    \includegraphics[width=0.5\textwidth]{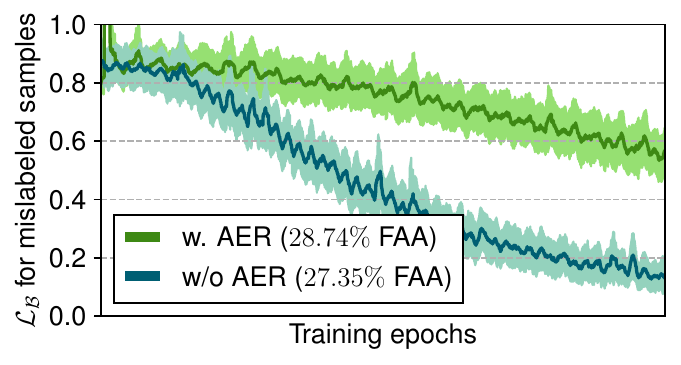}
    \caption{Effect of AER on the speed at which the model learns the noisy data}
    \label{fig:loss_stream}
\end{figure}
Here, we further provide evidence of the impact of AER on the overall performance of the model. In particular, in~\cref{fig:loss_stream} we depict the final accuracy (FAA) and the loss of the noisy samples from the current task of ER-ACE with and without AER during the second task of Split CIFAR-10. 

Surprisingly, we find that AER vastly reduces the rate of convergence of noisy samples, which just by itself improves over the baseline in terms of FAA. Indeed, in rehearsal CNL providing a purified and diverse set of examples to counter forgetting is only part of the challenge: as the model is subjected to a continuous stream of noisy data from the current task, an important effect is to reduce the speed with which noisy samples from the present are learned.
\section{On the computational demands of CLN methods}
We perform all the experiments on a Tesla V100-SXM2-16GB GPU. In~\cref{tab:timecompute} we report the computational costs of different methods in our setting, in terms of runtime and consumed memory.

Not only our method achieves superior performance, as shown in~\cref{table:ss_faa_new}, but also exhibits a lower overall training time.
\begin{table}[t]
\begin{center}
\begin{tabular}{lccc}
\hline
\noalign{\smallskip}

\textbf{Computational Cost} & \textbf{PuriDivER} & \textbf{PuriDivER.ME} & \textbf{OURs} \\
\noalign{\smallskip}

\hline
\noalign{\smallskip}

Total Time (\textit{hours}) & 5h15m & 2h50m & \textbf{1h30m} \\
Epoch Time (\textit{seconds}) & 76.90s & \textbf{15.69s} & 18.56s \\
Task Time (\textit{minutes}) & 73m50s & 19m39s & \textbf{16m33s} \\
Memory Used (\textit{GB}) & 7.77 & 7.50  & \textbf{6.75} \\
\hline
\end{tabular}
\end{center}
\caption{Comparison of Computational Cost $[\downarrow]$ of different methods when trained on CIFAR-10 with 40\% noise.}
\label{tab:timecompute}

\end{table}
\section{Additional details on SPR and CNLL}
In the main paper we provide a comparison between our proposal and SPR and CNLL. Nevertheless, as these methods were originally designed for the single-epoch setting, we had to design specific adjustments to make them viable for our scenario.

SPR initially stores samples in a \textit{delayed buffer} -- then splits into clean and noisy sets, with the former stored in a separate long-term buffer -- and then optimizes the model for approximately 7,000 training iterations through a self-supervised (SSL) objective. This implies that SPR involves approximately $448\times$ iterations than standard training\footnote{assuming a buffer size of 500, batch size of 32, and 10,000 samples}, making it unfeasible for our scenario due to time constraints. Indeed, while on CIFAR-10 our method takes around 16 minutes to complete 1 task (\cref{tab:timecompute}), SPR would require over 119 hours. We thus opt to distribute the training iterations of SPR across 25 epochs (see~\cref{tab:sprcnll}). Finally, as SPR employs two distinct memory buffers, we set the buffer size to 1000 for a fair comparison.

CNLL uses variable-length buffers to store confident clean and noisy samples, which implies a CL setting with unrestricted memory across tasks. To ensure fairness in comparison, we adhere to the well-established memory-budgeted CL~\cite{chaudhry2019tiny,van2019three,buzzega2020dark} setting. Thus, for CNLL we allocate a total memory budget of 2,500 exemplars across all 5 buffers specified by the original method.

As we move from the single to multi-epoch setting, we find a reduced effectiveness of the regularization of CNLL; such a result is in line with our hypothesis of~\cref{sec:aer}: as more epochs are allowed to learn the current task, sample selection based on the small-loss criterion fails to distinguish clean and noisy samples. Moreover, we find that such an outcome is maintained even in an unrestricted setting, where the memory budget is not a concern.

Finally, the performance gap w.r.t.\ our proposal is even more pronounced for SPR\textsuperscript{\textdaggerdbl}, where our method attains significantly higher accuracy in considerably less time; indeed, our reduced version of SPR requires around $109\times$ more time than our proposal, in line with our estimation.
\section{Additional details on the experimental settings and Final Forgetting}
To evaluate our proposal we build upon the open-source codebase provided by Mammoth~\cite{buzzega2020dark,caccia2022new,boschini2022class}, a CL framework based on \texttt{PyTorch}.

\subsection*{On the choice of datasets and noise}
\label{sec:datasets}
We empirically validate our method on four different classification benchmarks as mentioned in the main paper. For experiments on CIFAR-10/100~\cite{krizhevsky2009learning} and NTU-60~\cite{shahroudy2016ntu}, we corrupt the labels of the datasets at hand to obtain different noise configurations, which we then keep fixed for each of the experiments for fairness of results comparison across multiple methods.

In the process of injecting symmetric noise, we replace the ground-truth label with probability $r \in [0, 1]$ determined by the designated noise rate. The asymmetric or \textit{class-dependent} noise setting is an approximation of real-world corruption patterns, which alters labels within the same superclass. For example, in the CIFAR-100 dataset, each image comes with a "fine" label (specific class) and a "coarse" label (superclass). Here, label transitions are parameterized by $r$ such that the wrong class and true class have probability $r$ and $1 - r$, respectively. This results in sample ambiguity occurring only between similar classes, as it would in a realistic scenario. 

In each experiment, samples from the main dataset are split into disjoint sets based on their class and organized into tasks, following the \classil setup. 
We obtain the following versions of the datasets.\\
\noindent\textbf{\textit{Seq.\ CIFAR-10}}~The original dataset contains $50,000$ train and $10,000$ test low-resolution color images in 10 different classes. During training the model encounters 2 classes per task, namely (\quotationmarks{airplane}, \quotationmarks{car}), (\quotationmarks{bird}, \quotationmarks{cat}), (\quotationmarks{deer}, \quotationmarks{dog}), (\quotationmarks{frog}, \quotationmarks{horse}), (\quotationmarks{ship}, \quotationmarks{truck}).\\
\noindent\textbf{\textit{Seq.\ CIFAR-100}}~This original dataset is like the CIFAR-10, except it has 100 classes with 600 images each. Images are grouped into 20 superclasses, thus each image comes with a "fine" label (the class to which it belongs) and a "coarse" label (the superclass to which it belongs). Following this categorization, we organize classes in 10 tasks, each containing 5 classes from the same superclass.\\
\noindent\textbf{\textit{Seq.\ NTU-60}}~It comprises 60 action classes with 56,880 video samples, including 3D skeletal data (25 body joints per frame), all captured simultaneously using three Kinect V2 cameras. We here split the dataset into 6 tasks of 10 classes each.\\
\noindent\textbf{\textit{Seq.\ Food-101N}}~The dataset is composed of 101 web-crawled food images, split into 5 tasks (the first 4 containing 20 classes and the latter containing the remaining 21). Each class is relatively balanced, with an average of around 523 images per class and a standard deviation of around 11 (totaling 52867 images resized to $224\times 224$). The dataset contains instance-level noise, thus simulating a real-world scenario.

Notice that since some labels are incorrect, real class distribution for each task might vary. Details on the noisy labels injected on Seq.\ CIFAR-10/100, Seq.\ NTU-60 are released with the code.

\subsection*{Training details}
\noindent{\textbf{\textit{Architecture}}}~We use ResNet~\cite{he2016deep} family as a backbone for all the methods involved in our evaluation. ResNet18 is used for CIFAR-10/100 and ResNet34 is used for Food-101N, as in~\cite{bang2022online}. All the experiments do not feature pretraining.

\noindent{\textbf{\textit{Augmentation}}}~We apply random crops and horizontal flips to both stream and buffer examples, for each dataset at hand. For the implementations of PuriDivER, we use AutoAugment~\cite{cubuk2019autoaugment} as in the original paper~\cite{bang2022online}. 

\noindent{\textbf{\textit{Training}}}~We deliberately hold batch size out of the hyperparameter space and keep it fixed to $32$ for both stream and buffer examples. For each task, we train for 50 epochs for CIFAR-10/100, and 20 for Food-101N.

\noindent{\textbf{\textit{Buffer consolidation with MixMatch}}}~At the end of each task, we finetune the model on the buffer examples only, for 255 epochs. During this stage, we use SGD with Warm Restart (SGDR) through Cosine Annealing and a batch size of 64. For the purpose of label co-refinement, we set the number of different augmentations $\eta$ of~\cref{equ:mixmatch} to perform on the samples in the \textit{uncertain} set to 3.
\begin{table*}[t!]
\begin{center}
\setlength\tabcolsep{4pt}
\begin{small}
\begin{tabular}{lccc|cc|cc}
\hline\noalign{\smallskip}
\textbf{Benchmark} & \multicolumn{5}{c}{\textbf{Seq.\ CIFAR-100}} & \multicolumn{2}{c}{\textbf{Seq.\ NTU-60}}\\
\noalign{\smallskip}
\hline
\noalign{\smallskip}
&\multicolumn{3}{c}{\textit{symm}}  &\multicolumn{2}{c}{\textit{asymm}} &\multicolumn{2}{c}{\textit{symm}}\\

\textbf{Noise rate} & 20 & 40 & 60 & 20 & 40 & 20 & 40 \\
\hline
\noalign{\smallskip}
Joint            &  {0}       &  {0}        &  {0}        &  {0}        &  {0}     &  {0} &  {0} \\
Finetune      &  {81.52}    &  {71.51}    &  {57.96}    &  {73.91}    &  {54.71}    &  {85.09} & {73.46} \\
\hline
\noalign{\smallskip}
Reservoir                            &  {55.88} &  {55.79} &  {44.90} &  {43.48} &  {35.22} &  {54.53} & {61.33} \\
\hspace{.3em} + \textit{CoTeaching}   &  {55.20} &  {54.95} &  {36.07} &  {54.77} &  {26.03} & {34.30} & {29.03} \\
\hspace{.3em} + \textit{DivideMix}  &  {22.33} &  {26.73} &  {20.94} &  {23.45} &  {16.73} &   {18.45} & {18.70} \\
\hline
\noalign{\smallskip}
PuriDivER                             &      {20.52} &  {18.21} &  {14.77} &  {22.51} &  {17.26} &   {41.29} & {34.25} \\ 
PuriDivER.ME\textsuperscript{\textbf{\dag}} & {24.34} &  {25.06} &  {26.83} &  {25.40} &  {21.82} &   {25.76} & {18.41} \\ 
\hline
\noalign{\smallskip}
\textbf{OURs}                             &  {22.89}             &  {21.26}              &  {22.13}              &  {21.19}              &  {16.90}                        & {12.94} & {14.05}\\
\hspace{.3em} \textit{w. consolidation} &  {\underline{19.03}} &  {\underline{11.67}}  &  {\underline{12.02}}  &  {\underline{20.15}}  &  {\underline{9.28}} & {\underline{8.54}}  & {\underline{0.29}}\\
\hline
\end{tabular}
\end{small}
\end{center}
\caption{Final Forgetting (FF) $[\downarrow]$ of CNL methods on our selection of bencharks. \textsuperscript{\textbf{\dag}} Additional baselines created by adapting existing loss-based and CL approaches to the multi-epoch scenario.}
\label{tab:forgetting}

\end{table*}

\subsection*{Results in terms of Final Forgetting}
We repeat each of the experiments five times. We report in~\cref{table:ss_faa_new} of the main paper the Final Average Accuracy for all the experiments, with standard error values.

We also provide the final forgetting measure in~\cref{eq:ff} for all methods of the main comparison in~\cref{tab:forgetting}.
\begin{equation}
\label{eq:ff}
FF \triangleq \frac{1}{T-1}\sum_{j=0}^{T-2}{f_j},  
 \text{s.t.}   f_j=\max_{t\in{0,\ldots,T-2}}{a_j^t-a_j^{T-1}}
\end{equation}where $a_j^t$ is the accuracy of the model on the $j^{th}$ task after training on $t$ tasks. These additional results depict a  \textbf{lower} degree of \textbf{forgetting} of our proposal w.r.t. the baselines. 

When paired with~\cref{table:ss_faa_new} of the manuscript, such evidence shows higher overall effectiveness in learning from a noisy source of data, allowing more stable convergence on the current task and lower losses due to forgetting.
\section{Hyperparameters}
We choose to use different buffer sizes relying on the dataset length. For experiments conducted on CIFAR-10 and CIFAR-100, the buffer size is set to 500 and 2000, respectively. We set the buffer size to 500 for experiments on NTU. Finally, we use a buffer size of 2000 for Food-101.

We select the other hyperparameters by performing a grid search and using the Final Average Accuracy (FAA) as the selection criterion for the best parameters.
Here, we report the best values for each model, categorized by dataset and noise type.
\subsubsection*{CIFAR-10}

Noise type: \textit{sym} -- 20\%
\begin{itemize}[leftmargin=*]
\setlength\itemsep{0em}
\item\textbf{Joint}: $lr$: 0.03
\item\textbf{SGD}: $lr$: 0.03
\item\textbf{OURs}: $lr$: 0.03
\item\textbf{OURs + consolidation}: $lr$: 0.03; $lr_{\text{consolidation}}$: 0.1; $\lambda_u$: 0.01
\item\textbf{ER}: $lr$: 0.1; $lr_{\text{buffer fit.}}$: 0.05
\item\textbf{ER + CoTeaching}: $lr$: 0.1; $lr_{\text{buffer fit.}}$: 0.05
\item\textbf{ER + DivideMix}: $lr$: 0.1; $lr_{\text{buffer fit.}}$: 0.05
\item\textbf{PuriDivER}: $lr$: 0.001; $lr_{\text{buffer fit.}}$: 0.05; $\alpha$: 0.1
\item\textbf{PuriDivER.ME}: $lr$: 0.03; $lr_{\text{buffer fit.}}$: 0.05; $\alpha$: 0.1
\item\textbf{GDumb}: $lr_{\text{buffer fit.}}$: 0.1
\item\textbf{GDumb + CoTeaching}: $lr_{\text{buffer fit.}}$: 0.01
\item\textbf{GDumb + DivideMix}: $lr_{\text{buffer fit.}}$: 0.03
\end{itemize}
Noise type: \textit{sym} -- 40\%
\begin{itemize}[leftmargin=*]
\setlength\itemsep{0em}
\item\textbf{Joint}: $lr$: 0.03
\item\textbf{SGD}: $lr$: 0.03
\item\textbf{OURs}: $lr$: 0.03
\item\textbf{OURs + consolidation}: $lr$: 0.03; $lr_{\text{consolidation}}$: 0.1; $\lambda_u$: 0.01
\item\textbf{ER}: $lr$: 0.1; $lr_{\text{buffer fit.}}$: 0.05
\item\textbf{ER + CoTeaching}: $lr$: 0.03; $lr_{\text{buffer fit.}}$: 0.05
\item\textbf{ER + DivideMix}: $lr$: 0.03; $lr_{\text{buffer fit.}}$: 0.05
\item\textbf{PuriDivER}: $lr$: 0.001; $lr_{\text{buffer fit.}}$: 0.05; $\alpha$: 0.1
\item\textbf{PuriDivER.ME}: $lr$: 0.03; $lr_{\text{buffer fit.}}$: 0.05; $\alpha$: 0.1
\item\textbf{GDumb}: $lr_{\text{buffer fit.}}$: 0.03
\item\textbf{GDumb + CoTeaching}: $lr_{\text{buffer fit.}}$: 0.03
\item\textbf{GDumb + DivideMix}: $lr_{\text{buffer fit.}}$: 0.03
\end{itemize}
Noise type: \textit{sym} -- 60\%
\begin{itemize}[leftmargin=*]
\setlength\itemsep{0em}
\item\textbf{Joint}: $lr$: 0.03
\item\textbf{SGD}: $lr$: 0.03
\item\textbf{OURs}: $lr$: 0.03
\item\textbf{OURs + consolidation}: $lr$: 0.03; $lr_{\text{consolidation}}$: 0.1; $\lambda_u$: 0.01
\item\textbf{ER}: $lr$: 0.1; $lr_{\text{buffer fit.}}$: 0.1
\item\textbf{ER + CoTeaching}: $lr$: 0.1; $lr_{\text{buffer fit.}}$: 0.05
\item\textbf{ER + DivideMix}: $lr$: 0.1; $lr_{\text{buffer fit.}}$: 0.05
\item\textbf{PuriDivER}: $lr$: 0.001; $lr_{\text{buffer fit.}}$: 0.05; $\alpha$: 0.1
\item\textbf{PuriDivER.ME}: $lr$: 0.03; $lr_{\text{buffer fit.}}$: 0.05; $\alpha$: 0.1
\item\textbf{GDumb}: $lr_{\text{buffer fit.}}$: 0.03
\item\textbf{GDumb + CoTeaching}: $lr_{\text{buffer fit.}}$: 0.03
\item\textbf{GDumb + DivideMix}: $lr_{\text{buffer fit.}}$: 0.01
\end{itemize}
\subsubsection*{CIFAR-100}

Noise type: \textit{sym} -- 20\%
\begin{itemize}[leftmargin=*]
\setlength\itemsep{0em}
\item\textbf{Joint}: $lr$: 0.03
\item\textbf{SGD}: $lr$: 0.03
\item\textbf{OURs}: $lr$: 0.03
\item\textbf{OURs + consolidation}: $lr$: 0.03; $lr_{\text{consolidation}}$: 0.05; $\lambda_u$: 0.01
\item\textbf{DividERMix}: $lr$: 0.03
\item\textbf{ER}: $lr$: 0.03; $lr_{\text{buffer fit.}}$: 0.05
\item\textbf{ER + CoTeaching}: $lr$: 0.03; $lr_{\text{buffer fit.}}$: 0.05
\item\textbf{ER + DivideMix}: $lr$: 0.1; $lr_{\text{buffer fit.}}$: 0.01
\item\textbf{PuriDivER}: $lr$: 0.001; $lr_{\text{buffer fit.}}$: 0.05; $\alpha$: 0.1
\item\textbf{PuriDivER.ME}: $lr$: 0.03; $lr_{\text{buffer fit.}}$: 0.05; $\alpha$: 0.1
\item\textbf{GDumb}: $lr_{\text{buffer fit.}}$: 0.05
\item\textbf{GDumb + CoTeaching}: $lr_{\text{buffer fit.}}$: 0.05
\item\textbf{GDumb + DivideMix}: $lr_{\text{buffer fit.}}$: 0.05
\end{itemize}
Noise type: \textit{sym} -- 40\%
\begin{itemize}[leftmargin=*]
\setlength\itemsep{0em}
\item\textbf{Joint}: $lr$: 0.03
\item\textbf{SGD}: $lr$: 0.03
\item\textbf{OURs}: $lr$: 0.03
\item\textbf{OURs + consolidation}: $lr$: 0.03; $lr_{\text{consolidation}}$: 0.1; $\lambda_u$: 0.1
\item\textbf{ER}: $lr$: 0.03; $lr_{\text{buffer fit.}}$: 0.05
\item\textbf{ER + CoTeaching}: $lr$: 0.03; $lr_{\text{buffer fit.}}$: 0.05
\item\textbf{ER + DivideMix}: $lr$: 0.1; $lr_{\text{buffer fit.}}$: 0.05
\item\textbf{PuriDivER}: $lr$: 0.001; $lr_{\text{buffer fit.}}$: 0.05; $\alpha$: 0.1
\item\textbf{PuriDivER.ME}: $lr$: 0.03; $lr_{\text{buffer fit.}}$: 0.05; $\alpha$: 0.1
\item\textbf{GDumb}: $lr_{\text{buffer fit.}}$: 0.05
\item\textbf{GDumb + CoTeaching}: $lr_{\text{buffer fit.}}$: 0.05
\item\textbf{GDumb + DivideMix}: $lr_{\text{buffer fit.}}$: 0.05
\end{itemize}
Noise type: \textit{sym} -- 60\%
\begin{itemize}[leftmargin=*]
\setlength\itemsep{0em}
\item\textbf{Joint}: $lr$: 0.03
\item\textbf{SGD}: $lr$: 0.03
\item\textbf{OURs}: $lr$: 0.03
\item\textbf{OURs + consolidation}: $lr$: 0.03; $lr_{\text{consolidation}}$: 0.1; $\lambda_u$: 0.1
\item\textbf{ER}: $lr$: 0.03; $lr_{\text{buffer fit.}}$: 0.05
\item\textbf{ER + CoTeaching}: $lr$: 0.03; $lr_{\text{buffer fit.}}$: 0.05
\item\textbf{ER + DivideMix}: $lr$: 0.03; $lr_{\text{buffer fit.}}$: 0.05
\item\textbf{PuriDivER}: $lr$: 0.001; $lr_{\text{buffer fit.}}$: 0.05; $\alpha$: 0.1
\item\textbf{PuriDivER.ME}: $lr$: 0.03; $lr_{\text{buffer fit.}}$: 0.05; $\alpha$: 0.1
\item\textbf{GDumb}: $lr_{\text{buffer fit.}}$: 0.05
\item\textbf{GDumb + CoTeaching}: $lr_{\text{buffer fit.}}$: 0.05
\item\textbf{GDumb + DivideMix}: $lr_{\text{buffer fit.}}$: 0.05
\end{itemize}
Noise type: \textit{asym} -- 20\%
\begin{itemize}[leftmargin=*]
\setlength\itemsep{0em}
\item\textbf{Joint}: $lr$: 0.03
\item\textbf{SGD}: $lr$: 0.03
\item\textbf{OURs}: $lr$: 0.03
\item\textbf{OURs + buffer fit}: $lr$: 0.03; $lr_{\text{buffer fit.}}$: 0.05
\item\textbf{OURs + consolidation}: $lr$: 0.03; $lr_{\text{consolidation}}$: 0.05; $\lambda_u$: 0.005
\item\textbf{ER}: $lr$: 0.03; $lr_{\text{buffer fit.}}$: 0.05
\item\textbf{ER + CoTeaching}: $lr$: 0.03; $lr_{\text{buffer fit.}}$: 0.05
\item\textbf{ER + DivideMix}: $lr$: 0.03; $lr_{\text{buffer fit.}}$: 0.01
\item\textbf{PuriDivER}: $lr$: 0.001; $lr_{\text{buffer fit.}}$: 0.05; $\alpha$: 0.1
\item\textbf{PuriDivER.ME}: $lr$: 0.03; $lr_{\text{buffer fit.}}$: 0.05; $\alpha$: 0.1
\item\textbf{GDumb}: $lr_{\text{buffer fit.}}$: 0.05
\item\textbf{GDumb + CoTeaching}: $lr_{\text{buffer fit.}}$: 0.05
\item\textbf{GDumb + DivideMix}: $lr_{\text{buffer fit.}}$: 0.1
\end{itemize}
Noise type: \textit{asym} -- 40\%
\begin{itemize}[leftmargin=*]
\setlength\itemsep{0em}
\item\textbf{Joint}: $lr$: 0.03
\item\textbf{SGD}: $lr$: 0.03
\item\textbf{OURs}: $lr$: 0.03
\item\textbf{OURs + consolidation}: $lr$: 0.1; $lr_{\text{consolidation}}$: 0.05; $\lambda_u$: 0.1
\item\textbf{ER}: $lr$: 0.03; $lr_{\text{buffer fit.}}$: 0.05
\item\textbf{ER + CoTeaching}: $lr$: 0.03; $lr_{\text{buffer fit.}}$: 0.05
\item\textbf{ER + DivideMix}: $lr$: 0.1; $lr_{\text{buffer fit.}}$: 0.01
\item\textbf{PuriDivER}: $lr$: 0.001; $lr_{\text{buffer fit.}}$: 0.05; $\alpha$: 0.1
\item\textbf{PuriDivER.ME}: $lr$: 0.03; $lr_{\text{buffer fit.}}$: 0.05; $\alpha$: 0.1
\item\textbf{GDumb}: $lr_{\text{buffer fit.}}$: 0.05
\item\textbf{GDumb + CoTeaching}: $lr_{\text{buffer fit.}}$: 0.05
\item\textbf{GDumb + DivideMix}: $lr_{\text{buffer fit.}}$: 0.1
\end{itemize}

\subsubsection*{NTU RGB-D}

Noise type: \textit{sym} -- 20\%
\begin{itemize}[leftmargin=*]
\setlength\itemsep{0em}
\item\textbf{Joint}: $lr$: 0.1
\item\textbf{SGD}: $lr$: 0.1
\item\textbf{OURs}: $lr$: 0.1
\item\textbf{OURs + consolidation}: $lr$: 0.1; $lr_{\text{buffer fit.}}$: 0.1; $\lambda_r$: 0.01
\item\textbf{DividERMix}: $lr$: 0.03
\item\textbf{ER}: $lr$: 0.03; $lr_{\text{buffer fit.}}$: 0.05
\item\textbf{ER + CoTeaching}: $lr$: 0.1; $lr_{\text{buffer fit.}}$: 0.05
\item\textbf{ER + DivideMix}: $lr$: 0.1; $lr_{\text{buffer fit.}}$: 0.05
\item\textbf{PuriDivER}: $lr$: 0.3; $lr_{\text{buffer fit.}}$: 0.05; $\alpha$: 0.1
\item\textbf{PuriDivER.ME}: $lr$: 0.03; $lr_{\text{buffer fit.}}$: 0.05; $\alpha$: 0.1
\item\textbf{GDumb}: $lr_{\text{buffer fit.}}$: 0.03
\item\textbf{GDumb + CoTeaching}: $lr_{\text{buffer fit.}}$: 0.1
\item\textbf{GDumb + DivideMix}: $lr_{\text{buffer fit.}}$: 0.3
\end{itemize}
Noise type: \textit{sym} -- 40\%
\begin{itemize}[leftmargin=*]
\setlength\itemsep{0em}
\item\textbf{Joint}: $lr$: 0.1
\item\textbf{SGD}: $lr$: 0.03
\item\textbf{OURs}: $lr$: 0.1
\item\textbf{OURs + consolidation}: $lr$: 0.1; $lr_{\text{buffer fit.}}$: 0.1; $\lambda_r$: 0.01
\item\textbf{ER}: $lr$: 0.03; $lr_{\text{buffer fit.}}$: 0.05
\item\textbf{ER + CoTeaching}: $lr$: 0.1; $lr_{\text{buffer fit.}}$: 0.05
\item\textbf{ER + DivideMix}: $lr$: 0.1; $lr_{\text{buffer fit.}}$: 0.05
\item\textbf{PuriDivER}: $lr$: 0.3; $lr_{\text{buffer fit.}}$: 0.05; $\alpha$: 0.1
\item\textbf{PuriDivER.ME}: $lr$: 0.03; $lr_{\text{buffer fit.}}$: 0.05; $\alpha$: 0.1
\item\textbf{GDumb}: $lr_{\text{buffer fit.}}$: 0.1
\item\textbf{GDumb + CoTeaching}: $lr_{\text{buffer fit.}}$: 0.1
\item\textbf{GDumb + DivideMix}: $lr_{\text{buffer fit.}}$: 0.03
\end{itemize}
\end{document}